\documentclass{article}
\usepackage{iclr2027_conference,times}
\def\iclrruler#1{}
\usepackage{float}
\usepackage{stfloats}
\usepackage{hyperref}
\usepackage{url}
\usepackage{graphicx}
\usepackage{booktabs}
\usepackage{multirow}
\usepackage{amsmath,amssymb,amsthm}
\usepackage{algorithm}
\usepackage{algpseudocode}
\usepackage{placeins}
\usepackage{array}
\usepackage{xspace}

\theoremstyle{definition}
\newtheorem{proposition}{Proposition}
\theoremstyle{plain}
\newcommand{\RBA}{\textsc{RBA}\xspace}

\newcommand{\Cpos}{C^{+}}
\newcommand{\Cmix}{C^{\pm}}
\newcommand{\Czero}{C^{0}}
\newcommand{\ConflictSFT}{Conflict-SFT\xspace}
\definecolor{RBArow}{RGB}{220,234,246}
\makeatletter
\newcommand{\RBArowshade}{%
  \noalign{%
    \color{RBArow}%
    \hrule height \dimexpr\ht\@arstrutbox+\dp\@arstrutbox\relax
    \vskip-\dimexpr\ht\@arstrutbox+\dp\@arstrutbox\relax}}
\makeatother

\title{Regime Boundary Alignment for Evidence-Gated Question Answering}

\author{
Zeyan Li\textsuperscript{1} \quad
Qirong Guo\textsuperscript{2} \quad
Siyuan Qiu\textsuperscript{1} \quad
Hu Xu\textsuperscript{1} \quad
Chun Li\textsuperscript{1} \quad
Jianfeng Xu\textsuperscript{1} \\
\\[4pt]
\textsuperscript{1}Shanghai Jiao Tong University \\
\textsuperscript{2}The Hong Kong University of Science and Technology (Guangzhou)
}

\iclrfinalcopy

\begin{document}
\maketitle

\begin{abstract}
Retrieval-augmented language models are expected to answer from the retrieved evidence, but in practice they often keep answering when that evidence is missing. We trace this behavior to the training signal: answer-focused fine-tuning assigns no target to unsupported contexts, so it cannot distinguish a reader that abstains from one that guesses, and unsupported answering stays near 100\% even as supported accuracy improves. We introduce Regime Boundary Alignment (RBA), which trains a single reader on matched variants of the same question and gold answer. The reader is trained to produce the gold answer when the context supports it, including when conflicting evidence is also present, and to abstain when the correct support is removed; inference is ordinary decoding, with no verifier, threshold, or regime label. On three multi-hop QA datasets across three seeds, RBA reduces the unsupported-answer rate by more than sixty percentage points relative to conflict-focused training while matching its supported accuracy. On a held-out TriviaQA retrieval-miss slice, the same reader reduces unsupported answering from 100\% to below 1\% while also improving supported accuracy. These results indicate that evidence-gated answering must be learned on both sides of the support boundary.
\end{abstract}

\section{Introduction}

Retrieval-augmented generation (RAG) is intended to ground answers in retrieved evidence \citep{lewis2020rag,guu2020realm,izacard-grave-2021-leveraging}. Retrieval, however, may omit the answer-bearing passage or surface related distractors. Under an evidence-grounded contract, a reader should not silently fill such gaps from memory. Parametric knowledge may help interpret the context, but it does not by itself license an answer.

Existing approaches address parts of this contract. Prompting and uncertainty estimation rely on instructions, confidence estimates, or tuned thresholds \citep{ren2023selfevaluation,kuhn2023semantic}; conflict-supervised training preserves a supported answer amid distractors; capability- and knowledge-aware post-training draws the decision boundary from estimated model knowledge \citep{franzmeyer2026halt,sun2025dividethenalign}; refusal-only supervision can minimize unsupported answering by refusing indiscriminately. What they share is that supervision attaches to answers: none of them specifies a target action for the case in which correct support is absent.
\begin{figure}[ht]
\centering
\includegraphics[width=\linewidth]{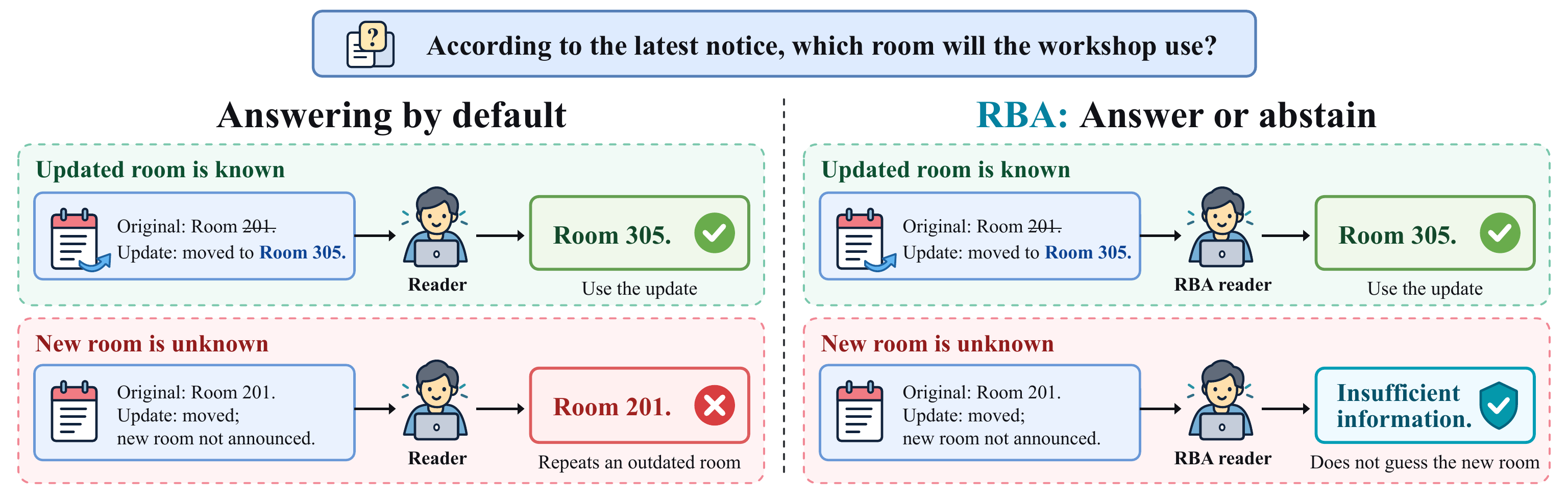}
\caption{The answerability boundary on one example. A notice that names the new room supports answering with it; a notice that confirms only the move does not, and repeating the old room is then unsupported. A reader trained only to answer (left) answers in both cases, while the \RBA{} reader (right) answers with support and abstains without it.}
\label{fig:teaser}
\end{figure}

The missing action defines a boundary. For a fixed question and gold answer, the reader should answer when the retrieved context contains sufficient correct support and abstain otherwise. Correct support plus a distractor remains answerable, since disagreement alone is not a reason to refuse (Figure~\ref{fig:teaser}). Standard answer-focused fine-tuning under-specifies this boundary: it rewards supported answering, and the same supported-answer loss is consistent with a reader that abstains on unsupported input and with a reader that guesses.

This gap is both formal and empirical. On finite observed regimes, answer-only likelihood leaves the no-support conditional unidentified, while a sequence loss with a matched abstention term controls a weighted upper bound on the pairwise answer--abstain boundary violations (Propositions 1 and 2). The empirical counterpart matches: on four held-out datasets, answer-focused conflict training reaches 66.13\% supported accuracy and still answers every no-support example.

We introduce Regime Boundary Alignment (\RBA), which trains a single reader on matched evidence conditions. For a fixed $(q,g)$, $\Cpos$ contains correct support, $\Cmix$ adds conflicting evidence without removing that support, and $\Czero$ removes the correct support while keeping the distractor. The reader learns the gold answer on $\Cpos$ and $\Cmix$ and a fixed abstention on $\Czero$ under one joint likelihood objective, and at inference it sees only the question and context, with no verifier, tuned threshold, or regime label.

Trained this way, the reader reduces the no-support unsupported-answer rate (UAR) from 100\% to 0\% while Supported Gold stays at 66.13\%; a paired bootstrap estimates the Supported Gold change at $-0.58$ percentage points with a 95\% interval of $[-1.84,0.67]$. Controls attribute the separation to the matched supervision rather than a generic refusal prior: replacing the query-aligned $\Czero$ with unrelated contexts leaves UAR at 66.58\%, and removing the answer anchor collapses Supported Gold to zero. The same $-100$-point UAR change repeats on Llama-3.1-8B and Mistral-7B readers, and the learned boundary transfers to dense-retrieval misses.

\paragraph{Contributions.}
\begin{enumerate}
    \item We formulate evidence-gated QA through a sequence-level answer--abstain boundary and construct matched retrieval regimes that vary support while holding the question and gold answer fixed.
    \item We introduce \RBA, a single joint likelihood objective, and show that it identifies the observed no-support action and upper-bounds pairwise boundary violations, without an inference-time verifier, threshold, or regime label.
    \item Across four datasets, three seeds, and three reader families, \RBA eliminates unsupported answering while preserving supported accuracy, and paired, component, and distribution-shift experiments attribute the effect to the matched two-sided supervision.
\end{enumerate}

\section{Related Work}

\paragraph{Evidence quality and acquisition.}
RAG and retrieval-augmented pretraining use external documents to supplement parametric knowledge \citep{lewis2020rag,guu2020realm}. One line of follow-up work measures how readers respond to imperfect evidence, covering irrelevant passages, evidence position, and retrieval noise \citep{yoran2024making,liu-etal-2024-lost,cuconasu2024power,shen-etal-2024-assessing}; RGB explicitly evaluates negative rejection alongside noise robustness, information integration, and counterfactual robustness \citep{chen2024benchmarking}, and RAGLens uses sparse-autoencoder features to detect unfaithful generations for post-hoc revision \citep{xiong2026raglens}. A second line modifies or extends the evidence itself: Self-RAG learns retrieval and critique actions \citep{asai2024selfrag}, CRAG evaluates retrieved evidence and triggers correction \citep{yan2024crag}, RECOMP compresses context and can omit unhelpful augmentation \citep{xu2024recomp}, Chain-of-Note assesses retrieved documents sequentially \citep{yu-etal-2024-chain}, and KnowGuard grounds clinical abstention in iterative knowledge-graph exploration across conversation turns \citep{dang2026knowguard}. Both lines treat the evidence as the moving part, whether to evaluate it, repair it, or extend it. \RBA{} instead holds the supplied context fixed and trains what the reader does with it: answer when the context supports the gold answer, and abstain when it does not.

\paragraph{Contextual and parametric knowledge.}
Controlled substitutions expose how readers arbitrate between retrieved statements and memorized facts \citep{longpre-etal-2021-entity,neeman-etal-2023-disentqa,xie2024adaptive}. Contrastive and adaptive decoding adjust their relative influence during generation \citep{shi-etal-2024-trusting,kim-etal-2024-adaptive,wang-etal-2025-adacad,yuan-etal-2024-discerning}. FaithfulRAG, Astute RAG, and MADAM-RAG resolve fact-level or document-level conflicts \citep{zhang-etal-2025-faithfulrag,wang-etal-2025-astute,wang2025retrievalaugmentedgenerationconflictingevidence}. Knowledgeable-R1 uses reinforcement learning to preserve parametric knowledge as a fallback under misleading retrieval \citep{lin2026knowledgeabler1}. These methods decide which source to trust when memory and evidence compete. \RBA{} studies the complementary case in which the required evidence is absent entirely: conflict with surviving support remains answerable, while a context without correct support receives an abstention target during training.

\paragraph{Abstention and selective prediction.}
Unanswerable QA and selective QA study withholding answers under missing evidence or distribution shift \citep{rajpurkar-etal-2018-know,kamath-etal-2020-selective}. LLM uncertainty methods estimate correctness through self-evaluation, verbalized confidence, or semantic uncertainty \citep{ren2023selfevaluation,lin2022teaching,kuhn2023semantic,kadavath2022language}; the Refusal Index measures whether refusal probability tracks error probability \citep{pan2026refusalindex}, and SAFER calibrates sampling and filtering to control miscoverage risk in open-ended QA \citep{wang2026safer}. Abstention can also be trained directly: DTA partitions examples by the availability of internal and retrieved knowledge and constructs quadrant-specific DPO preferences, including abstention when neither source suffices \citep{sun2025dividethenalign}, and HALT constructs capability-aligned fine-tuning responses by removing incorrect factual fragments or inserting uncertainty markers \citep{franzmeyer2026halt}. \RBA{} differs from the first group in that abstention is a generated response learned in training, not a post-hoc threshold on confidence; it differs from the second in how the action is defined---the question and gold answer are held fixed, answers are supervised under supported and mixed contexts, and abstention is supervised on the matched context with correct support removed.

\section{Problem Formulation}
\label{sec:setup}

\subsection{Evidence-grounded action boundary}

Let $q$ be a question, $g=(g_1,\ldots,g_{|g|})$ a gold answer sequence, and $C=\{d_1,\ldots,d_k\}$ the retrieved context. We reserve $a_0=\text{``Insufficient information.''}$ as the abstention sequence and assume $g\neq a_0$ by construction. An autoregressive reader assigns every candidate sequence $a$ the score
\begin{equation}
S_\theta(a\mid q,C)
=\sum_{t=1}^{|a|+1}\log p_\theta(a_t\mid q,C,a_{<t}),
\qquad a_{|a|+1}=\mathrm{EOS}.
\label{eq:sequence-score}
\end{equation}
Thus $S_\theta$ is a full sequence log probability, not a first-token score. The answer--abstain log-odds for a labeled pair $(q,g,C)$ are
\begin{equation}
m_\theta(q,g,C)
=S_\theta(g\mid q,C)-S_\theta(a_0\mid q,C)
=\log\frac{p_\theta(g\mid q,C)}{p_\theta(a_0\mid q,C)}.
\label{eq:boundary-margin}
\end{equation}
A positive margin prefers the gold answer to the fixed abstention; a negative margin reverses that preference. This margin is an analytical device that uses $g$ during training and analysis; it is not computed as an inference-time gate.

Let $s(C,g)\in\{0,1\}$ indicate whether $C$ contains sufficient correct support for $g$ under the task contract. The desired evidence-gated action and the corresponding pairwise sign condition are
\begin{equation}
a^\star(q,g,C)=
\begin{cases}
g, & s(C,g)=1,\\
a_0, & s(C,g)=0,
\end{cases}
\qquad
(2s(C,g)-1)m_\theta(q,g,C)>0.
\label{eq:policy}
\end{equation}
Here support concerns the supplied evidence, not whether the reader can recall $g$ parametrically. Parametric knowledge may help interpret a passage, but it does not license an unsupported answer in this setting.

\subsection{Matched context interventions}

For a fixed $(q,g)$, let $S$ denote answer-bearing support documents and $D$ denote query-related distractors. We construct one matched triplet $z=(q,g,\Cpos,\Cmix,\Czero)$ through
\begin{equation}
\Cpos=S,\qquad
\Cmix=S\cup D,\qquad
\Czero=D.
\label{eq:regimes}
\end{equation}
Consequently, $s(\Cpos,g)=s(\Cmix,g)=1$ and $s(\Czero,g)=0$. The contrast $\Cpos\!\rightarrow\!\Cmix$ adds conflict without removing evidence, whereas $\Cmix\!\rightarrow\!\Czero$ removes the answer-bearing support while retaining the distractor. Question identity, answer identity, and distractor source are held fixed. This matching prevents the target action from being explained solely by different question populations. The three desired inequalities can be summarized by the triplet boundary clearance
\begin{equation}
B_\theta(z)=
\min\!\left\{
m_\theta(q,g,\Cpos),\
m_\theta(q,g,\Cmix),\
-m_\theta(q,g,\Czero)
\right\}.
\label{eq:clearance}
\end{equation}
$B_\theta(z)>0$ means that one reader places this observed triplet on the correct sides of the answerability boundary.

Both $\Cpos$ and $\Cmix$ target $g$: the clean context preserves ordinary evidence-conditioned answering, and the mixed context prevents disagreement alone from becoming a refusal cue. Only $\Czero$ targets $a_0$. We instantiate these triplets from 2WikiMultihopQA, HotpotQA, MuSiQue, and TriviaQA \citep{ho-etal-2020-constructing,yang-etal-2018-hotpotqa,trivedi2022musiquemultihopquestionssinglehop,joshi-etal-2017-triviaqa}. Benchmark-annotated supporting documents form $\Cpos$; adding the paired adversarial documents yields $\Cmix$; retaining those distractor documents while removing all annotated support yields $\Czero$. We discard rows for which either side cannot be constructed. Because the three contexts share $(q,g)$ and the same distractor source, the main change is the availability of correct support rather than the question distribution.

\section{Regime Boundary Alignment}
\label{sec:method}

Section~\ref{sec:setup} leaves the no-support action without a training signal. \RBA supplies it directly, fitting the three target actions of Section~\ref{sec:setup} with one conditional generator: answering is anchored on both supported regimes and abstention on the matched no-support regime, with no separate classifier, router, or verifier.

\subsection{Sequence-level joint objective}

For target $a$, define the teacher-forced sequence loss
\begin{equation}
\ell_\theta(a;q,C)
=-S_\theta(a\mid q,C)
=-\sum_{t=1}^{|a|+1}\log p_\theta(a_t\mid q,C,a_{<t}).
\label{eq:sequence-nll}
\end{equation}
Prompt tokens are masked and EOS is included. We sum rather than length-average target-token NLLs.

For one matched triplet, \RBA minimizes
\begin{equation}
\mathcal{L}_{\RBA}(\theta;z)
=
\underbrace{\ell_\theta(g;q,\Cpos)
+\alpha\,\ell_\theta(g;q,\Cmix)}_{\mathcal{L}_{\mathrm{emit}}}
+\underbrace{\gamma\,\ell_\theta(a_0;q,\Czero)}_{\mathcal{L}_{\mathrm{withhold}}}.
\label{eq:rba}
\end{equation}
The clean term preserves ordinary evidence-conditioned answering. The mixed term preserves that action in the presence of conflict, so conflict alone does not become a refusal cue. The withhold term supplies the otherwise missing action at the matched retrieval failure. The main setting uses $\alpha=1$ and $\gamma=0.5$.

For training triplets $\mathcal{D}=\{z_i\}_{i=1}^{N}$, including the appended hard-replay triplets, the optimized empirical risk is
\begin{equation}
\widehat{\mathcal{R}}_{\RBA}(\theta)
=\frac{1}{N}\sum_{i=1}^{N}\mathcal{L}_{\RBA}(\theta;z_i).
\label{eq:empirical-risk}
\end{equation}
Hard replay changes which triplets occur in $\mathcal{D}$, not the objective. The decisive \ConflictSFT baseline is exactly the same answer anchor with the withhold coefficient removed:
\begin{equation}
\mathcal{L}_{\mathrm{conflict}}(\theta;z)
=\mathcal{L}_{\RBA}(\theta;z)\big|_{\gamma=0}
=\mathcal{L}_{\mathrm{emit}}.
\label{eq:conflict-objective}
\end{equation}
The RBA--\ConflictSFT comparison therefore isolates whether observing the no-support action changes the boundary while retaining the same two answer targets.

\subsection{Identification and boundary control}

\begin{proposition}
\label{prop:identification}
On a finite set of observed triplets, suppose targets are deterministic, no identical input is assigned conflicting targets, and the conditional sequence distributions are unrestricted at each observed input. Minimizing $\sum_i\mathcal{L}_{\mathrm{emit}}(\theta;z_i)$ leaves every observed $p_\theta(\cdot\mid q,\Czero)$ unidentified. For $\alpha,\gamma>0$, the pointwise minimizer of $\sum_i\mathcal{L}_{\RBA}(\theta;z_i)$ assigns unit mass to $g$ on $\Cpos,\Cmix$ and to $a_0$ on $\Czero$.
\end{proposition}

\begin{proof}
$\sum_i\mathcal{L}_{\mathrm{emit}}(\theta;z_i)$ contains no term evaluated at $\Czero$, so two conditional distributions that agree on $\Cpos$ and $\Cmix$ have identical loss regardless of their $\Czero$ actions. In the summed RBA objective, every observed regime has a positive-weight negative log likelihood. Over the probability simplex, each term is minimized by assigning unit mass to its deterministic target.
\end{proof}

Proposition~\ref{prop:identification} is an identification statement, not a claim that a finite neural optimizer reaches the simplex optimum. The next result connects the implemented sequence loss to the boundary signs in Equation~\ref{eq:clearance}. Let $\mathcal{K}=\{+,\pm,0\}$ index the regimes, with $y_+=y_{\pm}=1$, $y_0=-1$ and weights $w_+=1,w_{\pm}=\alpha,w_0=\gamma$. Write $m_r$ for the corresponding margin from Equation~\ref{eq:boundary-margin} and define the weighted violation count
\begin{equation}
V_\theta(z)
=\sum_{r\in\mathcal{K}}w_r\,\mathbf{1}\!\left[y_r m_r\leq 0\right].
\label{eq:violations}
\end{equation}

\begin{proposition}
\label{prop:surrogate}
For any autoregressive sequence distribution satisfying $p_\theta(g\mid q,C_r)>0$ and $p_\theta(a_0\mid q,C_r)>0$ for every $r\in\mathcal{K}$, and any $\alpha,\gamma\geq0$,
\begin{equation}
V_\theta(z)
\leq \frac{\mathcal{L}_{\RBA}(\theta;z)}{\log 2}.
\label{eq:surrogate-bound}
\end{equation}
\end{proposition}

\begin{proof}
For regime $r$, normalize the model probabilities of only $g$ and $a_0$. The target's binary negative log probability is $\log(1+\exp(-y_rm_r))$. It is no larger than the full sequence NLL because $p_\theta(g\mid q,C_r)+p_\theta(a_0\mid q,C_r)\leq1$. Whenever $y_rm_r\leq0$, this binary loss is at least $\log 2$. Multiplying by $w_r$ and summing gives Equation~\ref{eq:surrogate-bound}.
\end{proof}

Thus each term has a distinct geometric job: raise the two supported margins and reverse the no-support margin. The bound shows that the implemented sequence NLL is a valid surrogate for the three observed pairwise boundary decisions.

\begin{algorithm}[ht]
\caption{One Regime Boundary Alignment update.}
\label{alg:rba}
\small
\begin{algorithmic}[1]
\Require Batch $\mathcal{B}$ of $(q,g,\Cpos,\Cmix,\Czero)$
\State $\mathcal{J}\gets 0$
\For{$(q,g,\Cpos,\Cmix,\Czero)\in\mathcal{B}$}
    \State $\ell_{+}\gets\ell_\theta(g;q,\Cpos)$;
    $\ell_{\pm}\gets\ell_\theta(g;q,\Cmix)$
    \State $\ell_{0}\gets\ell_\theta(a_0;q,\Czero)$
    \State $\mathcal{J}\gets\mathcal{J}+\ell_{+}+\alpha\ell_{\pm}
    +\gamma\ell_{0}$
\EndFor
\State $\theta\gets\operatorname{OptimizerStep}(\theta,
\nabla_\theta\mathcal{J}/|\mathcal{B}|)$
\end{algorithmic}
\end{algorithm}

\paragraph{Why the no-support context is query-aligned.}
For triplet $i$, compare the aligned and shuffled constructions
\begin{equation}
\Czero_i=D_i,
\qquad
\widetilde{\Czero}_i=D_{\sigma(i)},\quad \sigma(i)\neq i.
\label{eq:query-alignment}
\end{equation}
RBA uses the distractors $D_i$ paired with $q_i$. The unaligned alternative instead uses a permutation across questions. Its loss constrains $p_\theta(a_0\mid q_i,\widetilde{\Czero}_i)$, not $p_\theta(a_0\mid q_i,\Czero_i)$. Without assumptions linking those inputs, low unaligned empirical risk does not identify the desired matched action. Query alignment therefore determines where the abstention anchor is placed; RQ2 tests whether parameter sharing transports a generic refusal association to the matched boundary.

\subsection{Training algorithm and complexity}

Algorithm~\ref{alg:rba} gives one optimization step. The three contexts for an example remain together in the same minibatch, so the answer and abstention signals update the same conditional generator under the same $(q,g)$. Sequence log probabilities are computed with teacher forcing and include the EOS token.

Let $n_{+}$, $n_{\pm}$, and $n_0$ denote the prompt-plus-target lengths for the three regimes, and let $T_\theta(n)$ be the cost of scoring a length-$n$ sequence. \RBA scores three sequences per triplet, for $O(T_\theta(n_{+})+T_\theta(n_{\pm})+T_\theta(n_0))$ work; concatenated along the batch dimension, this is arithmetic work rather than three serial model calls. For a dense $L$-layer Transformer of width $d$, $T_\theta(n)=O(L(n^2d+nd^2))$; \RBA changes only the constant number of scored sequences, not this asymptotic dependence. With LoRA rank $r$, trainable parameters scale as $O(r\sum_j(d_{j,\mathrm{in}}+d_{j,\mathrm{out}}))$ over adapted matrices. At inference, \RBA performs one ordinary reader generation and adds no retrieval pass, verifier, or routing model.

\begingroup
\setlength{\intextsep}{4pt}
\setlength{\abovecaptionskip}{5pt}
\begin{table}[htbp]
\centering
\scriptsize
\setlength{\tabcolsep}{2.6pt}
\renewcommand{\arraystretch}{1.08}
\caption{\textbf{Main result and mechanism controls.} Each dataset cell and
Macro report Supported Gold/UAR (G/U); abstention F1, AURC, and C@10 pool paired
supported/no-support rows. Values are seed medians and percentages except
AURC. Base and Prompt are deterministic single-checkpoint evaluations; all
trained rows use three seeds. The best displayed value within each individual
metric is bold (lower is better for UAR and AURC); displayed ties are included.}
\label{tab:main-results}
\resizebox{\textwidth}{!}{%
\begin{tabular}{lrrrrrrrr}
\toprule
Training condition & 2Wiki G/U & HotpotQA G/U & MuSiQue G/U & TriviaQA G/U &
Macro G/U & Abst.F1$\uparrow$ & AURC$\downarrow$ & C@10$\uparrow$\\
\midrule
\multicolumn{9}{l}{\textit{Reference and answer-focused baselines}}\\
Base & 0.78/100.00 & 2.03/100.00 & 0.00/100.00 & 8.59/100.00 &
2.85/100.00 & 0.00 & .980 & 0.00\\
Prompt & 27.34/8.59 & 46.70/17.26 & 9.09/7.14 & 37.11/99.61 &
30.06/33.15 & 63.24 & .304 & 11.36\\
DTA & 27.73/100.00 & 45.69/100.00 & 12.34/100.00 & 49.61/100.00 &
33.74/100.00 & 0.00 & .707 & 0.00\\
\ConflictSFT (no $\Czero$) & \textbf{72.27}/100.00 &
\textbf{64.97}/100.00 & 70.78/100.00 & 55.47/100.00 &
\textbf{66.13}/100.00 & 0.00 & .486 & 1.33\\
\midrule
\multicolumn{9}{l}{\textit{Boundary-supervision controls}}\\
No answer anchor & 0.00/\textbf{0.00} & 0.00/\textbf{0.00} &
0.00/\textbf{0.00} & 0.00/\textbf{0.00} &
0.00/\textbf{0.00} & 66.67 & .680 & 0.00\\
Unaligned $\Czero$ & \textbf{72.27}/57.81 & 63.45/58.88 & 70.13/50.00 &
\textbf{55.86}/99.22 & 65.62/66.58 & 47.13 & .284 & 16.86\\
No hard replay & 70.70/\textbf{0.00} & 63.96/\textbf{0.00} &
69.48/\textbf{0.00} & 52.34/\textbf{0.00} &
64.64/\textbf{0.00} & 99.83 & .039 & 86.73\\
\RBArowshade
\RBA & 71.09/\textbf{0.00} & 63.45/\textbf{0.00} &
\textbf{71.43}/\textbf{0.00} & 54.30/\textbf{0.00} &
\textbf{66.13}/\textbf{0.00} & \textbf{100.00} & \textbf{.037} &
\textbf{86.96}\\
\bottomrule
\end{tabular}}
\end{table}

\endgroup
\FloatBarrier

\section{Experiments}
\label{sec:experiments}

\subsection{Experimental setup}

\paragraph{Registered questions and data.}
The frozen campaign asks whether \RBA changes the boundary (RQ1), which
supervision is load-bearing (RQ2), where it generalizes (RQ3), and whether the
direction repeats across readers (RQ4). Training uses 2,000 natural BM25
triplets plus 147 hard TriviaQA triplets, with no evaluation-question overlap.
The paired test set contains 256 2Wiki, 197 HotpotQA, 154 MuSiQue, and 256
TriviaQA questions, each with supported and BM25 no-support views. RQ3 freezes
the RQ1 checkpoints and adds paired dense-retrieval views (95, 157, 232, and
256 questions for the four datasets), 256 partial-evidence examples per
multi-hop dataset, and 1,239 controlled seen, unseen-template, and
semantic-negation examples. Partial contexts retain one supporting document
but lack the remaining multi-hop chain.

\paragraph{Models and comparisons.}
The primary reader is Qwen3-8B in non-thinking mode. Base is unadapted, Prompt
adds an abstention instruction, \ConflictSFT uses
Equation~\ref{eq:conflict-objective}, and DTA is a matched-capacity
quadrant-preference baseline \citep{sun2025dividethenalign}. RQ2 removes or
misaligns individual RBA components; RQ4 repeats the decisive contrast with
Llama-3.1-8B-Instruct and Mistral-7B-Instruct-v0.3. All readers are rank-16 LoRA adapters (scale 32, dropout 0.05) trained for six epochs with AdamW at learning rate $10^{-4}$, with an effective batch size of 8 over 4 ranks. Training sequences are left-truncated at 3{,}072 tokens, and padded positions carry zero loss weight in every compared objective, so all rows supervise the same tokens.

\paragraph{Evaluation and aggregation.}
Generation is greedy with the training prompt, 3,000 input tokens, and at most
48 new tokens. Fixed lexical rules identify abstentions. Supported Gold is
normalized exact match; UAR is the answered fraction of no-support or partial
examples. Abstention F1 treats no-support as positive. Selective confidence is
the generated-sequence geometric-mean probability; lower AURC and higher
10\%-risk coverage are better. We macro-average datasets within each seed, then report median and range over seeds 42, 137, and 271; Base and Prompt are deterministic. F1, AURC, and
coverage instead pool matched positive/negative rows within seed. The decisive
contrast uses 20,000 question-stratified paired-bootstrap resamples, averaging
over seeds and macro-averaging datasets.

\subsection{RQ1: Learning the support boundary}

Table~\ref{tab:main-results} shows the per-dataset consistency and joint
trade-off. \ConflictSFT reaches 66.13\% macro Gold but answers every retrieval
miss; \RBA retains 66.13\% Gold and answers none of them. The UAR change from
100\% to 0\% occurs on every dataset, and the minimum \RBA abstention F1 over
seeds is 99.88\%. The identical macro medians hide a small per-dataset
pattern: \RBA gains 0.65 Gold points on MuSiQue and gives up between 1.17 and
1.52 on the other three datasets, which is why the medians coincide while the
paired bootstrap below estimates a small negative mean shift. Pooled F1 rises
from 0\% to 100\%, AURC falls from .486 to .037, and 10\%-risk coverage rises
from 1.33\% to 86.96\%. No other evaluated row keeps supported accuracy while
abstaining on retrieval misses.

The paired bootstrap quantifies the accuracy side of this trade-off. Across
the four datasets, \RBA minus \ConflictSFT is $-0.58$ percentage points in
supported Gold (95\% interval $[-1.84, 0.67]$) and $-100$ points in UAR
(interval $[-100, -100]$): supported accuracy is statistically
indistinguishable between the two readers, while the unsupported-answer rates
are completely separated.

The remaining rows fail in instructively different ways. Prompting lowers UAR
on the three multi-hop datasets (7.14--17.26\%) but barely moves it on
TriviaQA (99.61\%), where questions are single-hop factoids and a memorized
answer is most available; the abstention instruction loses exactly where the
model has an answer in memory. DTA fails on both axes at once: it answers
every no-support example and reaches only 33.74\% macro Gold, so
quadrant-specific preferences neither fit the abstention action nor preserve
reading accuracy in this setting. The no-answer-anchor row marks the opposite
extreme, and its abstention F1 of 66.67\% is a reminder that F1 alone does not
certify the boundary: on balanced paired views, a reader that abstains on
everything attains two-thirds F1, which is why we read F1 together with Gold
and UAR. Finally, the selective metrics show that the learned boundary is
reflected in confidence, not only in the emitted action. \ConflictSFT's
sequence confidence separates supported from no-support inputs only weakly
(.486 AURC, 1.33\% coverage at 10\% risk), while \RBA's separates them almost
perfectly (.037 AURC, 86.96\% coverage), so a practitioner could threshold
\RBA's confidence and recover most of the risk--coverage trade-off without
retraining.

\subsection{RQ2: Which supervision is load-bearing?}

The lower block of Table~\ref{tab:main-results} exposes opposite degenerate
policies under the two one-sided objectives. No-answer-anchor training always
abstains, producing 0\% Gold and 0\% UAR; removing $\Czero$ always answers,
producing 66.13\% Gold and 100\% UAR. Both targets are therefore necessary to
identify the desired action boundary.

The context attached to the abstention target matters as well. Replacing
query-aligned $\Czero$ with a random wrong document preserves 65.62\% Gold but
leaves UAR at 66.58\%: the reader learns a generic refusal association, not
the question-conditioned boundary. The failure is again uneven across
datasets: unaligned UAR stays between 50.00\% and 58.88\% on the three
multi-hop datasets but reaches 99.22\% on TriviaQA, the same dataset where
prompting fails. An instruction and a misaligned refusal target thus break
down at the same place, where the question most strongly invites a memorized
answer. Removing the 147 hard TriviaQA replay examples keeps median UAR at
0\% (at most 0.26\% across seeds) and lowers Gold to 64.64\%, so hard replay
contributes about 1.5 points of supported accuracy but does not carry the
abstention effect. The drop concentrates where the replay triplets come from:
TriviaQA loses 1.96 Gold points and MuSiQue 1.95, while HotpotQA is unchanged
within half a point.

\begin{figure}[ht]
\centering
\includegraphics[width=\linewidth]{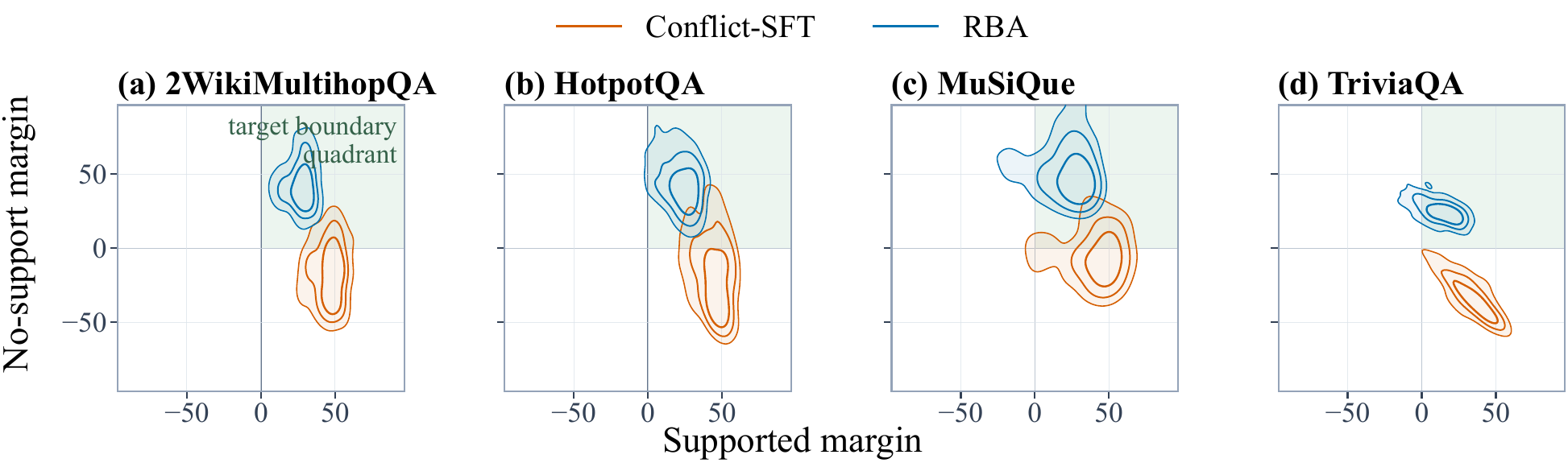}
\caption{\textbf{Evidence-boundary margin distributions.} Each contour is a
Gaussian KDE over held-out $(q,g)$ pairs after averaging exact teacher-forced
sequence margins within question across the three frozen seeds. The horizontal
axis is $S(g\mid q,\Cpos)-S(\mathrm{IDK}\mid q,\Cpos)$; the vertical axis is
$S(\mathrm{IDK}\mid q,\Czero)-S(g\mid q,\Czero)$. The shaded upper-right
quadrant therefore represents the desired joint preference.}
\label{fig:boundary-contours}
\end{figure}

\paragraph{Boundary geometry.}
Figure~\ref{fig:boundary-contours} scores the fixed gold and abstention
sequences on the paired held-out views, using the per-question margin of
Equation~\ref{eq:boundary-margin} averaged over the three frozen seeds.
\ConflictSFT's supported margin is positive, so it does prefer the gold answer
when support is present; its no-support margin, however, is predominantly
negative, meaning the reader still prefers the gold string over abstention
after support has been removed. Unsupported answering under \ConflictSFT is
therefore an active preference learned from a signal that never scored the
alternative, not a near-tie resolved by surface priors. \RBA shifts the
density into the positive-positive quadrant on all four datasets, and because
both margins are per question, the quadrant mass reflects jointly correct
decisions rather than two separately correct averages.

\begin{figure}[!htbp]
\centering
\includegraphics[width=0.90\linewidth]{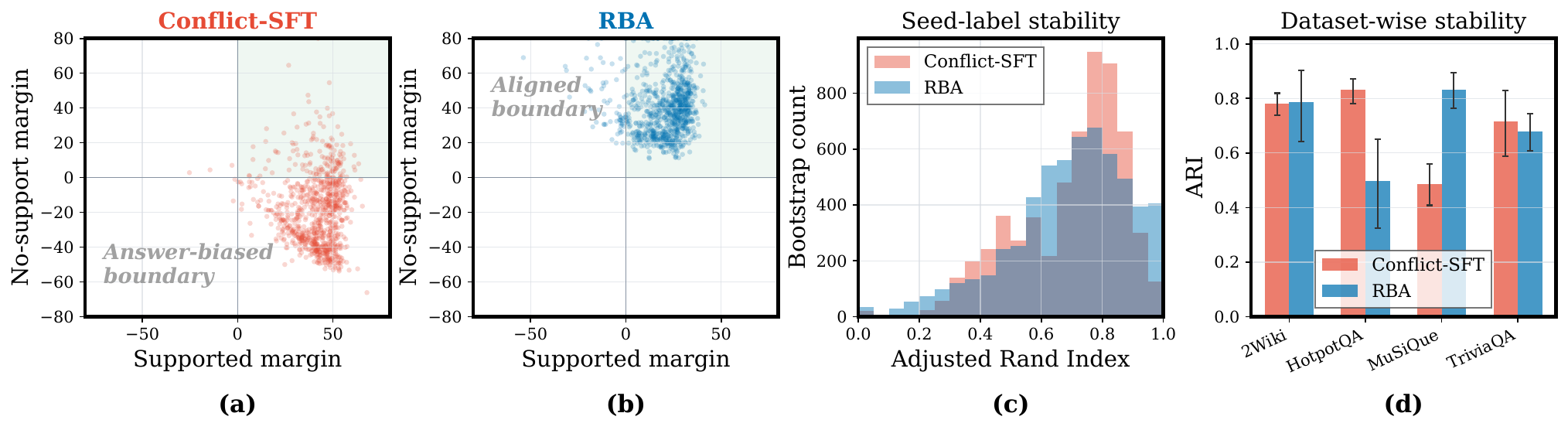}
\caption{\textbf{Boundary-state geometry and seed stability.} Panels (a--b)
show seed-averaged paired margin states for \ConflictSFT and \RBA. Panels
(c--d) compute adjusted Rand index (ARI) between boundary-quadrant labelings
of the same held-out questions across seed pairs, with bootstrap resampling
for the histogram and interquartile bars.}
\label{fig:boundary-cluster-stability}
\end{figure}
\FloatBarrier

\paragraph{Seed stability.}
Figure~\ref{fig:boundary-cluster-stability} checks that these boundaries are
properties of the trained readers rather than of a lucky seed: the quadrant
labelings of the same held-out questions are stable across seed pairs for both
readers, overall and per dataset. \ConflictSFT's answer-biased boundary is as
stable as \RBA's evidence-gated one, so the two readers differ in where the
boundary sits, not in whether training converges.

\subsection{RQ3: Where does the boundary generalize?}

\begin{table}[ht]
\centering
\small
\setlength{\tabcolsep}{4.0pt}
\caption{RQ3 frozen-checkpoint transfer. Entries are dataset-macro percentages
and seed medians. Paired rows report G/U; no-support-only rows report U.
Controlled conditions cover three multi-hop datasets.}
\label{tab:generalization}
\begin{tabular}{llrrr}
\toprule
Evidence shift & Metric & Base & \ConflictSFT & \RBA\\
\midrule
BM25 miss & G / U & 2.85 / 100.00 & 66.13 / 100.00 &
\textbf{66.13 / 0.00}\\
Dense miss & G / U & 3.12 / 100.00 & 61.70 / 100.00 &
\textbf{62.10 / 0.10}\\
Partial evidence & U & 100.00 & 100.00 & \textbf{83.07}\\
Controlled seen & U & 100.00 & 100.00 & \textbf{97.62}\\
Unseen template & U & 100.00 & 100.00 & \textbf{76.50}\\
Semantic negation & U & 100.00 & 100.00 & \textbf{99.30}\\
\bottomrule
\end{tabular}
\end{table}

The boundary transfers across retrieval algorithms: dense-miss UAR is 0.10\%
for \RBA versus 100\% for \ConflictSFT, with Gold at 62.10\% versus 61.70\%.
Because the dense views come from a different retriever, this transfer
indicates that the reader keys on the absence of support rather than on
surface properties of BM25 retrieval failures. It does not transfer broadly
across semantic forms of insufficiency. \RBA UAR is 83.07\% with partial
evidence and 97.62\%, 76.50\%, and 99.30\% on controlled seen,
unseen-template, and negation shifts. All four conditions stay far above the
BM25-miss level: the reader answers most semantic-insufficiency variants
regardless of phrasing, and partial evidence --- one genuine supporting
document with the rest of the chain missing --- is answered 83.07\% of the
time even though the full chain is absent. The unseen-template condition shows
the widest seed dispersion of any RQ3 row (66.25\% to 95.36\%), so what
partial transfer exists there is real but unstable. The learned boundary is
therefore robust to the retrieval algorithm but specific to retrieval misses:
partially supported and unseen phrasings of insufficiency are still answered a
substantial fraction of the time.

\subsection{RQ4: Does the direction repeat across readers?}

\begin{table}[htbp]
\centering
\small
\setlength{\tabcolsep}{3.4pt}
\caption{RQ4 cross-family comparison. Values are dataset-macro percentages and
medians over three seeds. Deltas are \RBA minus \ConflictSFT.}
\label{tab:xmodel}
\begin{tabular}{lrrrrrr}
\toprule
& \multicolumn{2}{c}{\ConflictSFT} & \multicolumn{2}{c}{\RBA} &
\multicolumn{2}{c}{Change}\\
\cmidrule(lr){2-3}\cmidrule(lr){4-5}\cmidrule(l){6-7}
Reader & G & U & G & U & $\Delta$G & $\Delta$U\\
\midrule
Qwen3-8B & 66.13 & 100.00 & 66.13 & 0.00 & +0.01 & -100.00\\
Llama-3.1-8B & 67.89 & 100.00 & 67.69 & 0.00 & -0.19 & -100.00\\
Mistral-7B & 66.76 & 100.00 & 66.71 & 0.00 & -0.05 & -100.00\\
\bottomrule
\end{tabular}
\end{table}

All three \ConflictSFT readers answer every no-support example; all \RBA
readers attain 0\% median UAR, with Gold changes within 0.2 points. Two
aspects of the table matter beyond the main effect. First, the Gold deltas are
small and centered on zero ($-0.19$, $-0.05$, $+0.01$), each well inside the
Qwen bootstrap interval of $[-1.84, 0.67]$ from RQ1: the cost of the boundary
does not grow across families within this size range. Second, the defect being
repaired is identical across families --- three readers with different
pretraining corpora and instruction tuning all arrive at a 100\% UAR boundary
after answer-focused training. That uniformity supports the paper's central
attribution: the missing action is a property of the training signal these
readers share, not of any one model family.

\FloatBarrier
\section{Conclusion}

We studied why retrieval-augmented readers keep answering when the retrieved
evidence does not support an answer. The cause is structural: training signals
that attach only to answers specify no action for unsupported contexts, so
unsupported answering is never penalized, regardless of how accurate the
reader becomes on supported questions.

\RBA closes this gap by training one reader on matched supported, conflicting,
and no-support views of each question, anchoring answering and abstaining in
the same likelihood. Across four datasets, three seeds, and three reader
families, the reader abstains on retrieval misses while its supported accuracy
remains statistically indistinguishable from answer-only conflict training.
Controls attribute the effect to the two-sided, question-aligned supervision
rather than to a generic refusal association or to the additional training
examples.

More broadly, the results suggest that evidence-gated behavior must be
supervised on both sides of the support boundary: a reader cannot learn when
not to answer from examples in which it always should. The boundary learned
here is specific to retrieval misses; extending it to semantic forms of
insufficiency, where the reader must judge whether partially relevant evidence
suffices, remains open.

\bibliography{references}

@inproceedings{shi-etal-2024-trusting,
    title = "Trusting Your Evidence: Hallucinate Less with Context-aware Decoding",
    author = "Shi, Weijia  and
      Han, Xiaochuang  and
      Lewis, Mike  and
      Tsvetkov, Yulia  and
      Zettlemoyer, Luke  and
      Yih, Wen-tau",
    editor = "Duh, Kevin  and
      Gomez, Helena  and
      Bethard, Steven",
    booktitle = "Proceedings of the 2024 Conference of the North American Chapter of the Association for Computational Linguistics: Human Language Technologies (Volume 2: Short Papers)",
    month = jun,
    year = "2024",
    address = "Mexico City, Mexico",
    publisher = "Association for Computational Linguistics",
    url = "https://aclanthology.org/2024.naacl-short.69/",
    doi = "10.18653/v1/2024.naacl-short.69",
    pages = "783--791"
}

@inproceedings{kim-etal-2024-adaptive,
    title = "Adaptive Contrastive Decoding in Retrieval-Augmented Generation for Handling Noisy Contexts",
    author = "Kim, Youna  and
      Kim, Hyuhng Joon  and
      Park, Cheonbok  and
      Park, Choonghyun  and
      Cho, Hyunsoo  and
      Kim, Junyeob  and
      Yoo, Kang Min  and
      Lee, Sang-goo  and
      Kim, Taeuk",
    editor = "Al-Onaizan, Yaser  and
      Bansal, Mohit  and
      Chen, Yun-Nung",
    booktitle = "Findings of the Association for Computational Linguistics: EMNLP 2024",
    month = nov,
    year = "2024",
    address = "Miami, Florida, USA",
    publisher = "Association for Computational Linguistics",
    url = "https://aclanthology.org/2024.findings-emnlp.136/",
    doi = "10.18653/v1/2024.findings-emnlp.136",
    pages = "2421--2431"
}

@inproceedings{wang-etal-2025-adacad,
    title = "{A}da{CAD}: Adaptively Decoding to Balance Conflicts between Contextual and Parametric Knowledge",
    author = "Wang, Han  and
      Prasad, Archiki  and
      Stengel-Eskin, Elias  and
      Bansal, Mohit",
    editor = "Chiruzzo, Luis  and
      Ritter, Alan  and
      Wang, Lu",
    booktitle = "Proceedings of the 2025 Conference of the Nations of the Americas Chapter of the Association for Computational Linguistics: Human Language Technologies (Volume 1: Long Papers)",
    month = apr,
    year = "2025",
    address = "Albuquerque, New Mexico",
    publisher = "Association for Computational Linguistics",
    url = "https://aclanthology.org/2025.naacl-long.581/",
    doi = "10.18653/v1/2025.naacl-long.581",
    pages = "11636--11652",
    ISBN = "979-8-89176-189-6"
}

@inproceedings{yuan-etal-2024-discerning,
    title = "Discerning and Resolving Knowledge Conflicts through Adaptive Decoding with Contextual Information-Entropy Constraint",
    author = "Yuan, Xiaowei  and
      Yang, Zhao  and
      Wang, Yequan  and
      Liu, Shengping  and
      Zhao, Jun  and
      Liu, Kang",
    editor = "Ku, Lun-Wei  and
      Martins, Andre  and
      Srikumar, Vivek",
    booktitle = "Findings of the Association for Computational Linguistics: ACL 2024",
    month = aug,
    year = "2024",
    address = "Bangkok, Thailand",
    publisher = "Association for Computational Linguistics",
    url = "https://aclanthology.org/2024.findings-acl.234/",
    doi = "10.18653/v1/2024.findings-acl.234",
    pages = "3903--3922"
}

@inproceedings{zhang-etal-2025-faithfulrag,
    title = "{F}aithful{RAG}: Fact-Level Conflict Modeling for Context-Faithful Retrieval-Augmented Generation",
    author = "Zhang, Qinggang  and
      Xiang, Zhishang  and
      Xiao, Yilin  and
      Wang, Le  and
      Li, Junhui  and
      Wang, Xinrun  and
      Su, Jinsong",
    editor = "Che, Wanxiang  and
      Nabende, Joyce  and
      Shutova, Ekaterina  and
      Pilehvar, Mohammad Taher",
    booktitle = "Proceedings of the 63rd Annual Meeting of the Association for Computational Linguistics (Volume 1: Long Papers)",
    month = jul,
    year = "2025",
    address = "Vienna, Austria",
    publisher = "Association for Computational Linguistics",
    url = "https://aclanthology.org/2025.acl-long.1062/",
    doi = "10.18653/v1/2025.acl-long.1062",
    pages = "21863--21882",
    ISBN = "979-8-89176-251-0"
}

@inproceedings{wang-etal-2025-astute,
    title = "Astute {RAG}: Overcoming Imperfect Retrieval Augmentation and Knowledge Conflicts for Large Language Models",
    author = "Wang, Fei  and
      Wan, Xingchen  and
      Sun, Ruoxi  and
      Chen, Jiefeng  and
      Arik, Sercan O",
    editor = "Che, Wanxiang  and
      Nabende, Joyce  and
      Shutova, Ekaterina  and
      Pilehvar, Mohammad Taher",
    booktitle = "Proceedings of the 63rd Annual Meeting of the Association for Computational Linguistics (Volume 1: Long Papers)",
    month = jul,
    year = "2025",
    address = "Vienna, Austria",
    publisher = "Association for Computational Linguistics",
    url = "https://aclanthology.org/2025.acl-long.1476/",
    doi = "10.18653/v1/2025.acl-long.1476",
    pages = "30553--30571",
    ISBN = "979-8-89176-251-0"
}

@inproceedings{wang2025retrievalaugmentedgenerationconflictingevidence,
      title={Retrieval-Augmented Generation with Conflicting Evidence},
      author={Wang, Han and Prasad, Archiki and Stengel-Eskin, Elias and Bansal, Mohit},
      booktitle={Second Conference on Language Modeling},
      year={2025},
      url={https://openreview.net/forum?id=z1MHB2m3V9},
}

@inproceedings{sun2025dividethenalign,
  title = {Divide-Then-Align: Honest Alignment based on the Knowledge Boundary of {RAG}},
  author = {Sun, Xin and Xie, Jianan and Chen, Zhongqi and Liu, Qiang and Wu, Shu and Chen, Yuehe and Song, Bowen and Wang, Zilei and Wang, Weiqiang and Wang, Liang},
  booktitle = {Proceedings of the 63rd Annual Meeting of the Association for Computational Linguistics (Volume 1: Long Papers)},
  year = {2025},
  month = jul,
  pages = {11461--11480},
  publisher = {Association for Computational Linguistics},
  address = {Vienna, Austria},
  doi = {10.18653/v1/2025.acl-long.561},
  url = {https://aclanthology.org/2025.acl-long.561/}
}

@inproceedings{yang-etal-2018-hotpotqa,
    title = "{H}otpot{QA}: A Dataset for Diverse, Explainable Multi-hop Question Answering",
    author = "Yang, Zhilin  and
      Qi, Peng  and
      Zhang, Saizheng  and
      Bengio, Yoshua  and
      Cohen, William  and
      Salakhutdinov, Ruslan  and
      Manning, Christopher D.",
    editor = "Riloff, Ellen  and
      Chiang, David  and
      Hockenmaier, Julia  and
      Tsujii, Jun{'}ichi",
    booktitle = "Proceedings of the 2018 Conference on Empirical Methods in Natural Language Processing",
    month = oct # "-" # nov,
    year = "2018",
    address = "Brussels, Belgium",
    publisher = "Association for Computational Linguistics",
    url = "https://aclanthology.org/D18-1259/",
    doi = "10.18653/v1/D18-1259",
    pages = "2369--2380"
}

@inproceedings{ho-etal-2020-constructing,
    title = "Constructing A Multi-hop {QA} Dataset for Comprehensive Evaluation of Reasoning Steps",
    author = "Ho, Xanh  and
      Duong Nguyen, Anh-Khoa  and
      Sugawara, Saku  and
      Aizawa, Akiko",
    editor = "Scott, Donia  and
      Bel, Nuria  and
      Zong, Chengqing",
    booktitle = "Proceedings of the 28th International Conference on Computational Linguistics",
    month = dec,
    year = "2020",
    address = "Barcelona, Spain (Online)",
    publisher = "International Committee on Computational Linguistics",
    url = "https://aclanthology.org/2020.coling-main.580/",
    doi = "10.18653/v1/2020.coling-main.580",
    pages = "6609--6625"
}

@article{trivedi2022musiquemultihopquestionssinglehop,
      title={MuSiQue: Multihop Questions via Single-hop Question Composition},
      author={Harsh Trivedi and Niranjan Balasubramanian and Tushar Khot and Ashish Sabharwal},
      journal={Transactions of the Association for Computational Linguistics},
      volume={10},
      year={2022},
      pages={539--554},
      address={Cambridge, MA},
      publisher={MIT Press},
      doi={10.1162/tacl_a_00475},
      url={https://aclanthology.org/2022.tacl-1.31/}
}

@inproceedings{joshi-etal-2017-triviaqa,
    title = "{T}rivia{QA}: A Large Scale Distantly Supervised Challenge Dataset for Reading Comprehension",
    author = "Joshi, Mandar  and
      Choi, Eunsol  and
      Weld, Daniel  and
      Zettlemoyer, Luke",
    editor = "Barzilay, Regina  and
      Kan, Min-Yen",
    booktitle = "Proceedings of the 55th Annual Meeting of the Association for Computational Linguistics (Volume 1: Long Papers)",
    month = jul,
    year = "2017",
    address = "Vancouver, Canada",
    publisher = "Association for Computational Linguistics",
    url = "https://aclanthology.org/P17-1147/",
    doi = "10.18653/v1/P17-1147",
    pages = "1601--1611"
}

@inproceedings{lewis2020rag,
  title = {Retrieval-Augmented Generation for Knowledge-Intensive {NLP} Tasks},
  author = {Lewis, Patrick and Perez, Ethan and Piktus, Aleksandra and Petroni, Fabio and Karpukhin, Vladimir and Goyal, Naman and K{\"u}ttler, Heinrich and Lewis, Mike and Yih, Wen-tau and Rockt{\"a}schel, Tim and Riedel, Sebastian and Kiela, Douwe},
  booktitle = {Advances in Neural Information Processing Systems},
  editor = {Larochelle, H. and Ranzato, M. and Hadsell, R. and Balcan, M. F. and Lin, H.},
  volume = {33},
  pages = {9459--9474},
  year = {2020},
  publisher = {Curran Associates, Inc.},
  url = {https://proceedings.neurips.cc/paper/2020/hash/6b493230205f780e1bc26945df7481e5-Abstract.html}
}

@inproceedings{guu2020realm,
  title = {Retrieval Augmented Language Model Pre-Training},
  author = {Guu, Kelvin and Lee, Kenton and Tung, Zora and Pasupat, Panupong and Chang, Mingwei},
  booktitle = {Proceedings of the 37th International Conference on Machine Learning},
  editor = {III, Hal Daum{\'e} and Singh, Aarti},
  series = {Proceedings of Machine Learning Research},
  volume = {119},
  pages = {3929--3938},
  year = {2020},
  month = {13--18 Jul},
  publisher = {PMLR},
  url = {https://proceedings.mlr.press/v119/guu20a.html}
}

@inproceedings{izacard-grave-2021-leveraging,
  title = {Leveraging Passage Retrieval with Generative Models for Open Domain Question Answering},
  author = {Izacard, Gautier and Grave, Edouard},
  booktitle = {Proceedings of the 16th Conference of the European Chapter of the Association for Computational Linguistics: Main Volume},
  editor = {Merlo, Paola and Tiedemann, Jorg and Tsarfaty, Reut},
  pages = {874--880},
  year = {2021},
  month = apr,
  address = {Online},
  publisher = {Association for Computational Linguistics},
  url = {https://aclanthology.org/2021.eacl-main.74/},
  doi = {10.18653/v1/2021.eacl-main.74}
}

@inproceedings{longpre-etal-2021-entity,
  title = {Entity-Based Knowledge Conflicts in Question Answering},
  author = {Longpre, Shayne and Perisetla, Kartik and Chen, Anthony and Ramesh, Nikhil and DuBois, Chris and Singh, Sameer},
  booktitle = {Proceedings of the 2021 Conference on Empirical Methods in Natural Language Processing},
  editor = {Moens, Marie-Francine and Huang, Xuanjing and Specia, Lucia and Yih, Scott Wen-tau},
  pages = {7052--7063},
  year = {2021},
  month = nov,
  address = {Online and Punta Cana, Dominican Republic},
  publisher = {Association for Computational Linguistics},
  url = {https://aclanthology.org/2021.emnlp-main.565/},
  doi = {10.18653/v1/2021.emnlp-main.565}
}

@inproceedings{neeman-etal-2023-disentqa,
  title = {{DisentQA}: Disentangling Parametric and Contextual Knowledge with Counterfactual Question Answering},
  author = {Neeman, Ella and Aharoni, Roee and Honovich, Or and Choshen, Leshem and Szpektor, Idan and Abend, Omri},
  booktitle = {Proceedings of the 61st Annual Meeting of the Association for Computational Linguistics (Volume 1: Long Papers)},
  editor = {Rogers, Anna and Boyd-Graber, Jordan and Okazaki, Naoaki},
  pages = {10056--10070},
  year = {2023},
  month = jul,
  address = {Toronto, Canada},
  publisher = {Association for Computational Linguistics},
  url = {https://aclanthology.org/2023.acl-long.559/},
  doi = {10.18653/v1/2023.acl-long.559}
}

@inproceedings{xie2024adaptive,
  title = {Adaptive Chameleon or Stubborn Sloth: Revealing the Behavior of Large Language Models in Knowledge Conflicts},
  author = {Xie, Jian and Zhang, Kai and Chen, Jiangjie and Lou, Renze and Su, Yu},
  booktitle = {The Twelfth International Conference on Learning Representations},
  year = {2024},
  url = {https://openreview.net/forum?id=auKAUJZMO6}
}

@inproceedings{yoran2024making,
  title = {Making Retrieval-Augmented Language Models Robust to Irrelevant Context},
  author = {Yoran, Ori and Wolfson, Tomer and Ram, Ori and Berant, Jonathan},
  booktitle = {The Twelfth International Conference on Learning Representations},
  year = {2024},
  url = {https://openreview.net/forum?id=ZS4m74kZpH}
}

@article{liu-etal-2024-lost,
  title = {Lost in the Middle: How Language Models Use Long Contexts},
  author = {Liu, Nelson F. and Lin, Kevin and Hewitt, John and Paranjape, Ashwin and Bevilacqua, Michele and Petroni, Fabio and Liang, Percy},
  journal = {Transactions of the Association for Computational Linguistics},
  volume = {12},
  pages = {157--173},
  year = {2024},
  address = {Cambridge, MA},
  publisher = {MIT Press},
  url = {https://aclanthology.org/2024.tacl-1.9/},
  doi = {10.1162/tacl_a_00638}
}

@inproceedings{cuconasu2024power,
  title = {The Power of Noise: Redefining Retrieval for {RAG} Systems},
  author = {Cuconasu, Florin and Trappolini, Giovanni and Siciliano, Federico and Filice, Simone and Campagnano, Cesare and Maarek, Yoelle and Tonellotto, Nicola and Silvestri, Fabrizio},
  booktitle = {Proceedings of the 47th International {ACM SIGIR} Conference on Research and Development in Information Retrieval},
  series = {{SIGIR} 2024},
  pages = {719--729},
  year = {2024},
  month = jul,
  publisher = {ACM},
  doi = {10.1145/3626772.3657834},
  url = {https://doi.org/10.1145/3626772.3657834}
}

@inproceedings{shen-etal-2024-assessing,
  title = {Assessing ``Implicit'' Retrieval Robustness of Large Language Models},
  author = {Shen, Xiaoyu and Blloshmi, Rexhina and Zhu, Dawei and Pei, Jiahuan and Zhang, Wei},
  booktitle = {Proceedings of the 2024 Conference on Empirical Methods in Natural Language Processing},
  editor = {Al-Onaizan, Yaser and Bansal, Mohit and Chen, Yun-Nung},
  pages = {8988--9003},
  year = {2024},
  month = nov,
  address = {Miami, Florida, USA},
  publisher = {Association for Computational Linguistics},
  url = {https://aclanthology.org/2024.emnlp-main.507/},
  doi = {10.18653/v1/2024.emnlp-main.507}
}

@inproceedings{asai2024selfrag,
  title = {{Self-RAG}: Learning to Retrieve, Generate, and Critique through Self-Reflection},
  author = {Asai, Akari and Wu, Zeqiu and Wang, Yizhong and Sil, Avirup and Hajishirzi, Hannaneh},
  booktitle = {The Twelfth International Conference on Learning Representations},
  year = {2024},
  url = {https://openreview.net/forum?id=hSyW5go0v8}
}

@misc{yan2024crag,
  title = {Corrective Retrieval Augmented Generation},
  author = {Yan, Shi-Qi and Gu, Jia-Chen and Zhu, Yun and Ling, Zhen-Hua},
  year = {2024},
  eprint = {2401.15884},
  archivePrefix = {arXiv},
  primaryClass = {cs.CL},
  doi = {10.48550/arXiv.2401.15884},
  url = {https://arxiv.org/abs/2401.15884}
}

@inproceedings{xu2024recomp,
  title = {{RECOMP}: Improving Retrieval-Augmented {LMs} with Context Compression and Selective Augmentation},
  author = {Xu, Fangyuan and Shi, Weijia and Choi, Eunsol},
  booktitle = {The Twelfth International Conference on Learning Representations},
  year = {2024},
  url = {https://openreview.net/forum?id=mlJLVigNHp}
}

@inproceedings{yu-etal-2024-chain,
  title = {Chain-of-Note: Enhancing Robustness in Retrieval-Augmented Language Models},
  author = {Yu, Wenhao and Zhang, Hongming and Pan, Xiaoman and Cao, Peixin and Ma, Kaixin and Li, Jian and Wang, Hongwei and Yu, Dong},
  booktitle = {Proceedings of the 2024 Conference on Empirical Methods in Natural Language Processing},
  pages = {14672--14685},
  year = {2024},
  month = nov,
  address = {Miami, Florida, USA},
  publisher = {Association for Computational Linguistics},
  url = {https://aclanthology.org/2024.emnlp-main.813/},
  doi = {10.18653/v1/2024.emnlp-main.813}
}

@inproceedings{kamath-etal-2020-selective,
  title = {Selective Question Answering under Domain Shift},
  author = {Kamath, Amita and Jia, Robin and Liang, Percy},
  booktitle = {Proceedings of the 58th Annual Meeting of the Association for Computational Linguistics},
  pages = {5684--5696},
  year = {2020},
  month = jul,
  address = {Online},
  publisher = {Association for Computational Linguistics},
  url = {https://aclanthology.org/2020.acl-main.503/},
  doi = {10.18653/v1/2020.acl-main.503}
}

@inproceedings{ren2023selfevaluation,
  title = {Self-Evaluation Improves Selective Generation in Large Language Models},
  author = {Ren, Jie and Zhao, Yao and Vu, Tu and Liu, Peter J. and Lakshminarayanan, Balaji},
  booktitle = {Proceedings on ``I Can't Believe It's Not Better: Failure Modes in the Age of Foundation Models'' at NeurIPS 2023 Workshops},
  editor = {Antor{\'a}n, Javier and Blaas, Arno and Buchanan, Kelly and Feng, Fan and Fortuin, Vincent and Ghalebikesabi, Sahra and Kriegler, Andreas and Mason, Ian and Rohde, David and Ruiz, Francisco J. R. and Uelwer, Tobias and Xie, Yubin and Yang, Rui},
  series = {Proceedings of Machine Learning Research},
  volume = {239},
  pages = {49--64},
  year = {2023},
  month = {16 Dec},
  publisher = {PMLR},
  url = {https://proceedings.mlr.press/v239/ren23a.html}
}

@inproceedings{kuhn2023semantic,
  title = {Semantic Uncertainty: Linguistic Invariances for Uncertainty Estimation in Natural Language Generation},
  author = {Kuhn, Lorenz and Gal, Yarin and Farquhar, Sebastian},
  booktitle = {The Eleventh International Conference on Learning Representations},
  year = {2023},
  eprint = {2302.09664},
  archivePrefix = {arXiv},
  primaryClass = {cs.CL},
  doi = {10.48550/arXiv.2302.09664},
  url = {https://arxiv.org/abs/2302.09664}
}

@article{lin2022teaching,
  title = {Teaching Models to Express Their Uncertainty in Words},
  author = {Lin, Stephanie and Hilton, Jacob and Evans, Owain},
  journal = {Transactions on Machine Learning Research},
  year = {2022},
  month = oct,
  url = {https://openreview.net/forum?id=8s8K2UZGTZ}
}

@misc{kadavath2022language,
  title = {Language Models (Mostly) Know What They Know},
  author = {Kadavath, Saurav and Conerly, Tom and Askell, Amanda and Henighan, Tom and Drain, Dawn and Perez, Ethan and Schiefer, Nicholas and Hatfield-Dodds, Zac and DasSarma, Nova and Tran-Johnson, Eli and Johnston, Scott and El-Showk, Sheer and Jones, Andy and Elhage, Nelson and Hume, Tristan and Chen, Anna and Bai, Yuntao and Bowman, Sam and Fort, Stanislav and Ganguli, Deep and Hernandez, Danny and Jacobson, Josh and Kernion, Jackson and Kravec, Shauna and Lovitt, Liane and Ndousse, Kamal and Olsson, Catherine and Ringer, Sam and Amodei, Dario and Brown, Tom and Clark, Jack and Joseph, Nicholas and Mann, Ben and McCandlish, Sam and Olah, Chris and Kaplan, Jared},
  year = {2022},
  eprint = {2207.05221},
  archivePrefix = {arXiv},
  primaryClass = {cs.CL},
  doi = {10.48550/arXiv.2207.05221},
  url = {https://arxiv.org/abs/2207.05221}
}

@inproceedings{rajpurkar-etal-2018-know,
  title = {Know What You Don{'}t Know: Unanswerable Questions for {SQ}u{AD}},
  author = {Rajpurkar, Pranav and Jia, Robin and Liang, Percy},
  booktitle = {Proceedings of the 56th Annual Meeting of the Association for Computational Linguistics (Volume 2: Short Papers)},
  pages = {784--789},
  year = {2018},
  month = jul,
  address = {Melbourne, Australia},
  publisher = {Association for Computational Linguistics},
  url = {https://aclanthology.org/P18-2124/},
  doi = {10.18653/v1/P18-2124}
}

@article{chen2024benchmarking,
  title = {Benchmarking Large Language Models in Retrieval-Augmented Generation},
  author = {Chen, Jiawei and Lin, Hongyu and Han, Xianpei and Sun, Le},
  journal = {Proceedings of the AAAI Conference on Artificial Intelligence},
  volume = {38},
  number = {16},
  pages = {17754--17762},
  year = {2024},
  doi = {10.1609/aaai.v38i16.29728},
  url = {https://ojs.aaai.org/index.php/AAAI/article/view/29728}
}

@inproceedings{franzmeyer2026halt,
  title = {High Accuracy, Less Talk ({HALT}): Reliable {LLMs} through Capability-Aligned Finetuning},
  author = {Franzmeyer, Tim and Sravankumar, Archie and Liu, Lijuan and Mao, Yuning and Hou, Rui and Wang, Sinong and Foerster, Jakob Nicolaus and Zettlemoyer, Luke and Khabsa, Madian},
  booktitle = {International Conference on Learning Representations},
  year = {2026},
  url = {https://openreview.net/forum?id=LYqBnNVaXD}
}

@inproceedings{dang2026knowguard,
  title = {{KnowGuard}: Knowledge-Driven Abstention for Multi-Round Clinical Reasoning},
  author = {Dang, Xilin and Chen, Kexin and Su, Xiaorui and Noori, Ayush and Arango, I{\~n}aki and Vittor, Lucas and Long, Xinyi and Du, Yuyang and Zitnik, Marinka and Heng, Pheng-Ann},
  booktitle = {International Conference on Learning Representations},
  year = {2026},
  url = {https://openreview.net/forum?id=gQRefH8upx}
}

@inproceedings{xiong2026raglens,
  title = {Toward Faithful Retrieval-Augmented Generation with Sparse Autoencoders},
  author = {Xiong, Guangzhi and He, Zhenghao and Liu, Bohan and Sinha, Sanchit and Zhang, Aidong},
  booktitle = {International Conference on Learning Representations},
  year = {2026},
  url = {https://openreview.net/forum?id=hgBZP67BkP}
}

@inproceedings{lin2026knowledgeabler1,
  title = {Resisting Contextual Interference in {RAG} via Parametric-Knowledge Reinforcement},
  author = {Lin, Chenyu and Wen, Yilin and Su, Du and Tan, Hexiang and Sun, Fei and Chen, Muhan and Bao, Chenfu and Lv, Zhonghou},
  booktitle = {International Conference on Learning Representations},
  year = {2026},
  url = {https://openreview.net/forum?id=6Qc6sO1jh9}
}

@inproceedings{pan2026refusalindex,
  title = {Can {LLMs} Refuse Questions They Do Not Know? Measuring Knowledge-Aware Refusal in Factual Tasks},
  author = {Pan, Wenbo and Xu, Jie and Chen, Qiguang and Dong, Junhao and Qin, Libo and Li, Xinfeng and Haining, Yu and Jia, Xiaohua},
  booktitle = {International Conference on Learning Representations},
  year = {2026},
  url = {https://openreview.net/forum?id=9gJBhkLRat}
}

@inproceedings{wang2026safer,
  title = {{SAFER}: Risk-Constrained Sample-then-Filter in Large Language Models},
  author = {Wang, Qingni and Fan, Yue and Wang, Xin Eric},
  booktitle = {International Conference on Learning Representations},
  year = {2026},
  url = {https://openreview.net/forum?id=kJmLmOvwLC}
}
\bibliographystyle{iclr2027_conference}

\end{document}